\documentstyle[12pt]{article}
\newtheorem{theorem}{Theorem}
\newtheorem{corollary}{Corollary}
\newtheorem{lemma}{Lemma}

\newtheorem{definition}{Definition}

\newcommand{\blackslug}{\mbox{\hskip 1pt \vrule width 4pt height 8pt 
depth 1.5pt \hskip 1pt}}
\newcommand{\QED}{\quad\blackslug\lower 8.5pt\null\par\noindent}
\newcommand{\proof}{\par\penalty-1000\vskip .5 pt\noindent{\bf Proof\/: }}

\newcommand{\cB}{\mbox{${\cal B}$}}
\newcommand{\cC}{\mbox{${\cal C}$}}

\newcommand{\cF}{\mbox{${\cal F}$}}

\newcommand{\cL}{\mbox{${\cal L}$}}

\newcommand{\cS}{\mbox{${\cal S}$}}

\newcommand{\Cn}{\mbox{${\cal C}n$}}

\newcommand{\Pf}{\mbox{${\cal P}_{f}$}}

\newcommand{\ra}{\rightarrow}
\newcommand{\Ra}{\Rightarrow}
\newcommand{\eqdef}{\stackrel{\rm def}{=}}

\newcommand{\subseteqf}{\mbox{$\subseteq_{f}$}}

\begin{document}
\bibliographystyle{plain}

\title{Deductive Nonmonotonic Inference Operations: Antitonic Representations
\thanks{
This work was 
partially supported 
by the Jean and Helene Alfassa fund for 
research in Artificial Intelligence}
}
\author{Yuri Kaluzhny 
\thanks{Mathematics Institute, Hebrew University, Jerusalem 91904 (Israel)}
\and Daniel Lehmann
\thanks{Institute of Computer Science, Hebrew University, 
Jerusalem 91904 (Israel)}
}

\maketitle

\begin{abstract}
We provide a characterization of those nonmonotonic
inference operations \cC\ for which $\cC(X)$ may be described as the set
of all logical consequences of $X$ together with some set of additional
assumptions $\cS(X)$ that depends anti-monotonically on $X$ (i.e., 
\mbox{$ X \subseteq Y $} implies \mbox{$ \cS(Y) \subseteq \cS(X) $}).
The operations represented are exactly characterized in terms of properties
most of which have been studied in~\cite{FL:IGPL}.
Similar characterizations of right-absorbing and cumulative operations
are also provided.
For cumulative operations, our results fit in closely with those 
of~\cite{Freund:supra93}.
We then discuss extending finitary operations to infinitary operations
in a canonical way and discuss co-compactness properties.
Our results provide a satisfactory notion of pseudo-compactness, generalizing
to deductive nonmonotonic operations the notion of compactness for monotonic
operations.
They also provide an alternative, more elegant and more general, 
proof of the existence
of an infinitary deductive extension for any finitary deductive operation
(Theorem~7.9 of~\cite{FL:IGPL}). 
\end{abstract}
\section{Introduction}
A fundamental intuition of {\em Default Reasoning} (understood 
in the wider sense)
is that a reasoner has, at its disposal, a set of facts $X$ 
(a fact is represented by a formula) and a set of defaults $D$
(there is no general agreement on the way the defaults ought to be 
represented). 
Given those, it draws conclusions (conclusions are formulas,
as facts are) by extracting from $D$ and $X$ some set of assumptions $A$
(formulas again), held in the presence of $X$, and then accepting as plausible
conclusions the set of logical consequences of the set
\mbox{$X \cup A$}.
Some nonmonotonic systems are explicitly presented in this way,
most notably the Closed World Assumption of Reiter's~\cite{Rei:CWA},
the system described by Poole in~\cite{Poole:88} (at least in its skeptical
version when the set of defaults is finite),
the system based on Epistemic Entrenchment~\cite{GarMak:92}, and
the Rational Closure construction of~\cite{LMAI:92}.
Other systems, most notably the Default Logic of~\cite{Reiter:80} 
and the Circumscription of~\cite{McCarthy:80} are presented, at first, 
in ways that
do not fit this paradigm, but they could have been presented this way:
Default Logic adds to the facts the conclusions of the applicable defaults,
and Circumscription adds to the facts what can be deduced from the defaults
when abnormalities are minimized.
In fact any system may be presented in this way by taking
for $A$ the set of plausible conclusions from $X$.

In such a presentation, it seems very natural to expect that, for any fixed set
of defaults $D$, the mapping \cS\ that sends a set of facts $X$ to the set
of assumptions \mbox{$A = \cS(X)$} held in the presence of $X$,
be antitonic (i.e., anti-monotonic,
i.e., \mbox{$ X \subseteq Y $} implies \mbox{$ \cS(Y) \subseteq \cS(X) $}).
Indeed, this is explicitly the case for the Closed World Assumption.
For finite Poole systems, even those without constraints, this is not
the case, though. There, given a set $D$ of formulas, $\cC(X)$ is defined as 
\mbox{$\Cn(X , \bigcap_{B \in \cB(X)} \Cn(B))$}, where $\cB(X)$ is the set
of all subsets of $D$ that are consistent with $X$ and maximal for this 
property. The intersection must not be antitonic.
Intuitively, antitonicity is a very natural property since 
$\cS(X)$ is some set of default assumptions
that are {\em compatible}, or consistent with $X$, and the larger $X$ is,
the less compatible (with $X$) formulas there are.
Antitonicity seems to be a necessary requirement for \cS, at least when
$X$ has a single {\em extension}. The purpose of this paper is to
characterize those operations that may be defined by such an antitonic
representation. From our characterization, will follow that many nonmonotonic
systems that were not originally presented in such a way are amenable 
to an antitonic presentation. This point will be taken up in the conclusion.
In particular finite Poole systems without constraints have an antitonic
representation, even though, as explained above, their natural presentation
is not antitonic.

In~\cite{Brass:93}, S. Brass considers a property (IMD, Definition 3.13)
that may seem closely related to the antitonicity of \cS.
In fact, IMD is very different from it. In our notations,
IMD may be described as: if \mbox{$x \in \cC(\emptyset)$}
and \mbox{$X \subseteq Y$}, then if $x$ is in $\cC(Y)$ it is also in
$\cC(X)$. There does not seem to be much intuitive support for such a property.

We suppose a language \cL\ is given, and, with it some 
consequence operation \Cn\ in the sense of Tarski.
The elements of \cL\ will be referred to as formulas.
About \Cn, we shall assume, as customary in the literature,
that \Cn\ satisfies inclusion, monotonicity, idempotency and compactness.
As usual, we write $\Cn(X,Y)$ instead of $\Cn(X \cup Y)$,
and \mbox{$X \models Y$} for \mbox{$Y \subseteq \Cn(X)$}.
We shall assume that the language \cL\ 
has {\em implication}. i.e.,
for any formulas \mbox{$a , b \in \cL$}, there is a formula 
\mbox{$a \ra b$} such that, for any \mbox{$X \subseteq \cL$}, 
\mbox{$b \in \Cn(X , a)$} iff \mbox{$a \ra b \in \Cn(X)$}.
Some of our results do not depend on this assumption, or could
be proved with weaker assumptions on \cL.
We shall use the following lemma from~\cite[Lemma A.3]{FL:IGPL}.
\begin{lemma}
\label{le:IGPL}
\ 
\begin{enumerate}
\item \label{arrow}
For any set $Y$ and any finite set $A$ of formulas,
there is a set \mbox{$A \ra Y$} of formulas such that,
for any set $X$ of formulas,
\mbox{$A \ra Y \subseteq \Cn(X)$} iff \mbox{$Y \subseteq \Cn(X , A)$}.
If $Y$ is finite, so is \mbox{$A \ra Y$}.
\item \label{adm}
For any \mbox{$X , Y ,  Z \subseteq \cL$},
\mbox{$\Cn(X , Y) \cap \Cn(X , Z) = \Cn(X , \Cn(Y) \cap \Cn(Z))$}.
\item \label{stradm}
For any finite set $A$ of formulas and any family 
\mbox{$Y_{i} , i \in I$} of sets of formulas,
\mbox{$\Cn(A , \bigcap_{i \in I} \Cn(Y_{i})) = $} \nolinebreak[3]
\mbox{$\bigcap_{i \in I} \Cn(A , Y_{i})$}.
\end{enumerate}
\end{lemma}
\section{Infinitary operations}
We consider infinitary operations
\mbox{$\cC : 2^{\,\cL} \longrightarrow 2^{\,\cL}$}.
We shall need to consider a host of properties for such operations.
Most of them were considered in~\cite{FL:IGPL}.
Our terminology is slightly different, since we consider operations that
are not always cumulative, or even absorbing.
After the definitions, we shall compare those properties of 
infinitary operations
with the corresponding properties of finitary consequence relations described
in~\cite{KLMAI:89}.

The following have to be understood 
for arbitrary subsets $X$ and $Y$ of \cL, and an arbitrary element $x$ of \cL.
We shall write \mbox{$A \subseteqf X$} if $A$ is a finite subset of $X$.
\begin{eqnarray}
{\rm (supraclassicality)} & \Cn(X) \subseteq \cC(X) \label{eq:supra} \\
{\rm (left \ absorption)} & \Cn(\cC(X)) = \cC(X) \label{eq:leftab} \\
{\rm (right \ absorption)} & \Cn(X) = \Cn(Y) \ \Ra \ \cC(X) = \cC(Y) 
\label{eq:rightab} \\
{\rm (deductivity)} & Y \subseteq X \ \Ra \ \cC(X) \subseteq 
\Cn( X , \cC(Y)) \label{eq:findeduc} \\
{\rm (cumulativity)} & Y \subseteq \Cn(\cC(X)) \ \Ra \ 
\Cn(\cC(X , Y)) = \Cn(\cC(X))
\label{eq:cum} \\
{\rm (antitonicity)} & X \subseteq Y \ \Ra \ \cC(Y) \subseteq \cC(X)
\label{eq:anti} \\
{\rm (compactness)} & x \in \cC(X) \ \Ra \ \exists A_{x} \subseteqf X 
\nonumber \\
& {\rm \ such \ that \ } \forall Y , {\rm \ if \ }
A_{x} \subseteqf Y \subseteq X , {\rm \ then \ } x \in \cC(Y)
\label{eq:comp} \\
{\rm (supracompactness)} & x \in \cC(X) \ \Ra \ \exists A_{x} \subseteqf X 
\nonumber \\
& {\rm \ such \ that \ } \forall Y , {\rm \ if \ }
A_{x} \subseteqf Y \subseteq \cC(X) , {\rm \ then \ } x \in \cC(Y)
\label{eq:supracomp}
\end{eqnarray}
Supraclassicality is a consequence of Reflexivity and Right-Weakening.
Left-absorption corresponds to Right-Weakening + And.
Right-absorption corresponds to Left-Logical-Equivalence.
The central property of this paper, Deductivity, corresponds to the
(S) rule. To understand its intuitive appeal, consider the case 
\mbox{$X = Y \cup \{ a \}$} ($a$ is an arbitrary formula).
Deductivity says that, if \mbox{$x \in \cC(Y , a)$}, then $x$ should already
be in
\mbox{$\Cn(a , \cC(Y))$}, i.e., that \mbox{$a \ra x$} should be
in \mbox{$\cC(Y)$}.
In other words, if $x$ is expected to be true on the evidence 
``$a$ and $Y$'', then, on the evidence $Y$ alone, we should already
expect ``if $a$ then $x$''.
Cumulativity corresponds to Cut + Cautious Monotonicity.

Right-absorption could have been written as \mbox{$\cC(X) = \cC(\Cn(X))$}.
We have chosen a formulation that generalizes without change to 
finitary operations.
Cumulativity is also defined in a slightly more convoluted way than
usual since we intend to consider cumulative operations that do not 
satisfy left-absorption.
The definition of supracompactness is taken from ~\cite{Freund:supra93}.
Our generalization of compactness is different both from the compactness
considered in~\cite{Freund:supra93}
and from supracompactnes.
It is clear that any supracompact operation is compact (our meaning), 
and that any compact operation in our sense is compact in Freund's sense.
For monotonic operations our notion of compactness coincides with
the usual one.
For supraclassical, left-absorbing, cumulative operations, we shall see
in Corollary~\ref{co:comp-supra} that compact operations are exactly
the supracompact operations of Freund.
Our first result characterizes those operations that have an antitonic
representation of the sort we discussed in the introduction.
The gist of our theorem is that,
assuming supraclassicality and left-absorption, the existence of an antitonic
representation is essentially equivalent to deductivity plus compacity.
Since compacity is of concern only for infinitary operations, for finitary
operations antitonic representations are equivalent to the (S) rule.
\begin{theorem}
\label{the:repinf}
Let \cC\ be an infinitary operation.
The following three properties are equivalent.
\begin{enumerate}
\item \label{antit} 
There is an antitonic operator \cS\ such that, for any set $X$
of formulas, \mbox{$\cC(X) = \Cn(X , \cS(X))$},
\item \label{inters}
for any set $X$,
\begin{equation}
\label{eq:inters}
\cC(X) = \Cn(X , \bigcap_{Y \subseteq X} \cC(Y)),
\end{equation}
\item \label{supra}
\cC\ is supraclassical, left-absorbing, deductive and compact.
\end{enumerate}
\end{theorem}
\proof
We show, first, that~\ref{antit} implies~\ref{supra}.
Let \mbox{$\cC(X) = \Cn(X ,\cS(X))$}, for some antitonic $\cS$.
The operation \cC\ is supraclassical because \Cn\ satisfies inclusion and
monotonicity.
It is left-absorbing because \Cn\ is idempotent.
Let us check it is deductive.
Suppose \mbox{$Y \subseteq X$}.
Then  \mbox{$\cS(X) \subseteq \cS(Y)$}, since \cS\ is antitonic, 
and by monotonicity of \Cn, 
\[
\cC(X) = \Cn(X , \cS(X)) \subseteq \Cn(X , \cS(Y)) \subseteq
\Cn(X , \Cn(Y , \cS(Y)))
= \Cn(X , \cC(Y)).
\]
Let us show it is also compact.
Suppose \mbox{$x \in \cC(X)$}. By the compactness of \Cn, there is a finite set
\mbox{$B_{x} \subseteqf X \cup \cS(X)$} such that 
\mbox{$x \in \Cn(B_{x}$}.
Let \mbox{$A_{x} \eqdef B_{x} \cap X$}.
We have
\mbox{$A_{x} \subseteqf X$}, and, by monotonicity of \Cn, 
\[
\Cn(B_{x}) = \Cn(A_{x} , B_{x} \cap \cS(X)) 
\subseteq \Cn(A_{x} , \cS(X)),
\]
and \mbox{$x \in \Cn(A_{x} , \cS(X))$}.
If \mbox{$A_{x} \subseteqf Y \subseteq X$}, then 
\mbox{$x \in \Cn(Y , \cS(Y))$}
by monotonicity of \Cn\ and antitonicity of \cS.

We now show that~\ref{supra} implies~\ref{inters}.
Suppose \cC\ is supraclassical, left-absorbing, deductive and compact.
Since, \mbox{$\bigcap_{Y \subseteq X} \cC(Y) \subseteq \cC(X)$}
and \cC\ is supraclassical and left-absorbing, we see easily that
\[
\Cn(X , \bigcap_{Y \subseteq X} \cC(Y)) \subseteq \cC(X).
\]
Suppose, now, that \mbox{$x \in \cC(X)$}.
Since \cC\ is compact, there is a finite set \mbox{$A \subseteqf X$},
such that \mbox{$x \in \cC(Y)$}, for any $Y$, 
\mbox{$A \subseteqf Y \subseteq X$}.
In other terms, \mbox{$x \in \cC(A , Y)$}, for any
\mbox{$Y \subseteq X$}.
Since \cC\ is deductive
\[
x \in \bigcap_{Y \subseteq X} \Cn(A , Y , \cC(Y)) = 
\bigcap_{Y \subseteq X} \Cn(A , \cC(Y)).
\]
By Lemma~\ref{le:IGPL}, part~\ref{stradm},
\[
x \in \Cn(A , \bigcap_{Y \subseteq X} \cC(Y)) \subseteq 
\Cn(X , \bigcap_{Y \subseteq X} \cC(Y)). 
\]
The fact that~\ref{inters} implies~\ref{antit} is obvious
since \mbox{$\cS(X) \eqdef \bigcap_{Y \subseteq X} \cC(Y)$} is
antitonic.
\QED
It is clear that there are supraclassical, left-absorbing 
deductive operations that are not compact.
Consider for example the operation \cC\ defined by 
\mbox{$\cC(X) = \Cn(X , a)$} if $X$ is infinite and 
\mbox{$\cC(X) = \Cn(X)$} otherwise.
The antitonic operator that appears in Equation~(\ref{eq:inters}),
i.e. \mbox{$\bigcap_{Y \subseteq X} \cC(Y)$}, is the largest
antitonic operator that represents \cC, as will be explained now.
\begin{lemma}
\label{le:large}
If, for some antitonic \cS, for any \mbox{$X \subseteq \cL$}, 
\mbox{$\cC(X) = \Cn(X , \cS(X))$} 
then, 
\mbox{$\cS(X) \subseteq \bigcap_{Y \subseteq X} \cC(Y)$}.
\end{lemma}
\proof
Let $Y$ be an arbitrary subset of $X$.
We have, since \cS\ is antitonic,
\[
\cS(X) \subseteq \cS(Y) \subseteq \Cn(Y , \cS(Y)) = \cC(Y).
\]
\QED
Our next result extends Theorem~\ref{the:repinf} to 
right-absorbing operations.
In~\cite{Freund:supra93} the intersection of Equation~\ref{eq:intersra} 
was called the trace of $X$.
\begin{theorem}
\label{the:repinfra}
Let \cC\ be an infinitary operation.
The following three properties are equivalent.
\begin{enumerate}
\item \label{antitra} 
There is an antitonic right-absorbing operator \cS\ such that, for any set $X$
of formulas, \mbox{$\cC(X) = \Cn(X , \cS(X))$},
\item \label{intersra}
for any set $X$,
\begin{equation}
\label{eq:intersra}
\cC(X) = \Cn(X , \bigcap_{X \models Y} \cC(Y)),
\end{equation}
\item \label{suprara}
\cC\ is supraclassical, left-absorbing, right-absorbing, 
deductive and compact.
\end{enumerate}
\end{theorem}
\proof
Let us show, first, that~\ref{antitra} implies~\ref{suprara}.
This is clear from Theorem~\ref{the:repinf} and the fact that, if \cS\
is right-absorbing, so is \mbox{$\Cn(X , \cS(X))$}.
We now show that~\ref{suprara} implies~\ref{intersra}.
From right-absorption of \cC\ and Theorem~\ref{the:repinf}, we have
\[
\cC(X) = \cC(\Cn(X)) = \Cn( \Cn(X) , \bigcap_{X \models Y} \cC(Y))
\] 
and we conclude easily.
The last leg of the proof is obvious since
\[
\cS(X) \eqdef \bigcap_{X \models Y} \cC(Y)
\] 
is antitonic and right-absorbing.
\QED
Notice that we do not claim that, for a right-absorbing \cC\ the 
intersections appearing in Equations~(\ref{eq:inters}) and~(\ref{eq:intersra})
are equal.
The following example will show it need not be the case.
Consider the propositional calculus on two variables $p$ and $q$, and let
\Cn\ be logical consequence. Define $\cS(X)$ as $\Cn(p)$ if 
\mbox{$\Cn(X) = \Cn(\emptyset)$} and $\emptyset$ otherwise.
The operator \cS\ is obviously antitonic and right-absorbing.
If \mbox{$\cC(X) \eqdef \Cn(X , \cS(X))$} one sees that
$p$ is an element of $\cC(\emptyset)$ and of $\cC(p)$, but not an element
of $\cC(q \ra p)$. Therefore
\mbox{$p \in \bigcap_{Y \subseteq \{ p \}} \cC(Y)$} but
\mbox{$p \not \in \bigcap_{p \models Y } \cC(Y)$}.
Similarly, to what we have shown in Lemma~\ref{le:large},
the antitonic operation, \mbox{$\bigcap_{X \models Y} \cC(Y)$}
is the largest right-absorbing antitonic representation of \cC.
\begin{lemma}
\label{le:largera}
If \mbox{$\cC(X) = \Cn(X , \cS(X))$} for some antitonic right-absorbing \cS,
then, \mbox{$\cS(X) \subseteq \bigcap_{X \models Y} \cC(Y)$}
\end{lemma}
\proof
Let $Y$ be an arbitrary subset of $\Cn(X)$.
We have, since \cS\ is right-absorbing and antitonic,
\[
\cS(X) = \cS(\Cn(X)) \subseteq \cS(Y) \subseteq \Cn(Y , \cS(Y)) = \cC(Y).
\]
\QED
We deal now with cumulative operations.
Our first result is important in itself and will be used in the proof of
our third characterization result.
Notice, first, that any supraclassical, left-absorbing cumulative operation
is right-absorbing, since
\mbox{$\Cn(X) \subseteq \cC(X)$} implies 
\mbox{$\Cn(\cC(X , \Cn(X)) = \Cn(\cC(X))$} and therefore
\mbox{$\cC(\Cn(X)) = \cC(X)$}.
\begin{theorem}
\label{the:cuminters}
If \cC\ is supraclassical, left-absorbing, deductive and cumulative,
then
\begin{equation}
\label{eq:cuminters}
\bigcap_{Y \subseteq \Cn(X)} \cC(Y) = \bigcap_{Y \subseteq \cC(X)} \cC(Y).
\end{equation}
\end{theorem}
\proof
Since \cC\ is supraclassical, the right-hand side is obviously a subset
of the left-hand side.
We must show that, for any \mbox{$Z \subseteq \cC(X)$},
\[
\bigcap_{Y \subseteq \Cn(X)} \cC(Y) \subseteq \cC(Z).
\]
But, since \mbox{$\Cn(X) \cap \Cn(Z) \subseteq \Cn(X)$},
\mbox{$\bigcap_{Y \subseteq \Cn(X)} \cC(Y) \subseteq \cC(\Cn(X) \cap \Cn(Z))$},
and it is enough to show that 
\mbox{$\cC(\Cn(X) \cap \Cn(Z)) \subseteq \cC(Z)$}.
We notice that, since \cC\ is left-absorbing and supraclassical,
\mbox{$\Cn(X) \cap \Cn(Z) \subseteq \cC(X)$}.
But \cC\ is left-absorbing and cumulative and we conclude that
\mbox{$\cC(X) = \cC(X , \Cn(X) \cap \Cn(Z))$} and therefore
\mbox{$Z \subseteq \cC(X , \Cn(X) \cap \Cn(Z))$}.
But \cC\ is deductive and
\[
\cC(X , \Cn(X) \cap \Cn(Z)) \subseteq 
\Cn(X , \Cn(X) \cap \Cn(Z) , \cC(\Cn(X) \cap \Cn(Z)))
\] 
and we conclude
that \mbox{$Z \subseteq \Cn(X , \cC(\Cn(X) \cap \Cn(Z)))$}.
But, obviously,
\[
Z \subseteq \Cn(Z , \cC(\Cn(X) \cap \Cn(Z))).
\]
By Lemma~\ref{le:IGPL}, part~\ref{adm},
\[
\Cn(X , \cC(\Cn(X) \cap \Cn(Z))) \cap \Cn(Z , \cC(\Cn(X) \cap \Cn(Z)))
\]
\[
= \Cn(X \cap Z , \cC(\Cn(X) \cap \Cn(Z))) = \cC(\Cn(X) \cap \Cn(Z)).
\]
The last equality follows from supraclassicality and left-absorption.
We conclude that
\mbox{$Z \subseteq \cC(\Cn(X) \cap \Cn(Z))$}.
By the cumulativity of \cC, we have
\mbox{$\cC(\Cn(X) \cap \Cn(Z)) = \cC(\Cn(X) \cap \Cn(Z) , Z)$}.
But \mbox{$\Cn(\Cn(X) \cap \Cn(Z) , Z) = \Cn(Z)$} and, since \cC\ is
right-absorbing by a remark above,
\mbox{$\cC(\Cn(X) \cap \Cn(Z)) = \cC(Z)$}.
\QED
We may now prove our third characterization theorem.
\begin{theorem}
\label{the:newrep}
Let \cC\ be an infinitary operation.
The following three properties are equivalent.
\begin{enumerate}
\item \label{antitcu}
There is an antitonic, right-absorbing and cumulative operator \cS\ such that,
for any set $X$ of formulas, \mbox{$\cC(X) = \Cn(X , \cS(X))$},
\item \label{interscumul}
for any set $X$,
\begin{equation}
\label{eq:interscumul}
\cC(X) = \Cn(X , \bigcap_{Y \subseteq \cC(X)} \cC(Y)),
\end{equation}
\item \label{supracu}
\cC\ is supraclassical, left-absorbing,
deductive, cumulative and compact.
\end{enumerate}
\end{theorem}
\proof
Let us show, first, that \ref{antitcu} implies \ref{supracu}.
Let \cC\ be as in~\ref{antitcu}. 
By Theorem~\ref{the:repinfra}, we only need to show
that \cC\ is cumulative.
Suppose, therefore, that we have 
\mbox{$Y \subseteq \Cn(\cC(X)) = \Cn(X , \cS(X))$}.
One of the inclusions we have to prove (the one corresponding to Cut)
is a consequence of the deductivity of \cC\ 
(using \mbox{$X \subseteq X \cup Y$}) and does not require
the cumulativity of \cS.
Indeed \mbox{$\cC(X , Y) \subseteq \Cn(X , Y , \cC(X))$} by the deductivity of
\cC. Since both $X$ and $Y$ are subsets of $\cC(X)$ and \cC\ is left-absorbing,
we conclude that \mbox{$\cC(X , Y) \subseteq \cC(X)$}.
The converse inclusion will be proved now.
Since \mbox{$Y \subseteq \Cn(X , \cS(X))$}, 
for any \mbox{$y \in Y$}, there exists a finite
\mbox{$A_{y} \subseteqf X$} such that \mbox{$y \in \Cn(A_{y} , \cS(X))$}.
We may apply part~\ref{arrow}
of Lemma~\ref{le:IGPL}, and consider the set 
\mbox{$W \eqdef \{ A_{y} \ra y \mid y \in Y \} \subseteq \Cn(\cS(X))$}.
But \cS\ is cumulative and therefore, 
\mbox{$\Cn(\cS(X)) = \Cn(\cS(X , W))$}.
One easily sees that \mbox{$\Cn(X , W) = \Cn(X , Y)$},
and, since \cS\ is right-absorbing, we have 
\linebreak[4]
\mbox{$\Cn(\cS(X)) = \Cn(\cS(X , Y))$}.
Therefore 
\[
\cC(X) = \Cn(X , \cS(X)) = \Cn(X , \cS(X , Y))
\subseteq \Cn(X , Y , \cS(X , Y)) = \cC(X , Y).
\]

Let us show now that \ref{supracu} implies \ref{interscumul}.
Suppose \cC\ is supraclassical, left-absorbing, deductive, cumulative
and compact.
As we noticed above, any supraclassical, left-absorbing and cumulative 
operation is right-absorbing.
We may use Theorem~\ref{the:repinfra} to see that 
Equation~\ref{eq:intersra} holds, and conclude by
Theorem~\ref{the:cuminters}.
Finally, let us show that \ref{interscumul} implies \ref{antitcu}.
Suppose \cC\ satisfies Equation~\ref{eq:interscumul}.
We shall show that
\mbox{$\cS(X) \eqdef \bigcap_{Y \subseteq \cC(X)} \cC(Y)$} is antitonic,
right-absorbing and cumulative.
None of those properties is obvious.
One immediately sees that \cC\ is supraclassical and left-absorbing.
One easily sees that \cC\ is deductive.
Indeed, suppose \mbox{$Y \subseteq X$}, by supraclassicality we have
\mbox{$Y \subseteq \cC(X)$} and therefore
\mbox{$\cS(X) \subseteq \cC(Y)$}.
We conclude that
\mbox{$\cC(X) = \Cn(X , \cS(X)) \subseteq \Cn(X , \cC(Y))$}.
The crux of the proof is to show that \cC\ is cumulative.
Let \mbox{$Z \subseteq \cC(X)$} (remember \cC\ is left-absorbing).
Since \cC\ is deductive, we have
\mbox{$\cC(X , Z) \subseteq \Cn(Z , \cC(X)) = \cC(X)$}.
But, in turn, this implies
\mbox{$\cS(X) \subseteq \cS(X , Z)$}, and
\[
\cC(X) = \Cn(X , \cS(X)) \subseteq \Cn(X , Z , \cS(X , Z)) = \cC(X , Z).
\]
We have shown that \cC\ is supraclassical, left-absorbing, deductive
and cumulative. By Theorem~\ref{the:cuminters}, we conclude that
\mbox{$\cS(X) = \bigcap_{X \models Y} \cC(Y)$} and we conclude
that \cS\ is antitonic and right-absorbing.
The only thing left to prove is that \cS\ is cumulative.
Suppose \mbox{$Z \subseteq \Cn(\cS(X))$}.
We see that \mbox{$Z \subseteq \cC(X)$}, and, since \cC\ is cumulative,
we have \mbox{$\cC(X) = \cC(X , Z)$}.
We conclude that \mbox{$\cS(X) = \cS(X , Z)$}.
\QED
The following corollary will make completely clear the relation
between compacity and supracompacity.
\begin{corollary}
\label{co:comp-supra}
Let \cC\ be supraclassical, left-absorbing, deductive and cumulative.
It is supracompact iff it is compact.
\end{corollary}
\proof
The {\em only if} part is obvious from the supraclassicality of \cC\ and
the definitions.
For the {\em if} part, by Theorem~\ref{the:newrep}, 
Equation~\ref{eq:interscumul} holds.
Supracompacity follows easily.
\QED
\section{Finitary deductive operations}
We shall denote the set of all finite subsets of $X$ by $\Pf(X)$.
We shall now consider finitary operations 
\mbox{$ \cF : \Pf(\cL) \longrightarrow 2^{\,\cL}$}.
The properties of supraclassicality, left-absorption, right-absorption, 
deductivity, antitonicity and cumulativity for finitary operations are
defined exactly as for infinitary operations, after replacing arbitrary
sets $X$ and $Y$ by finite sets $A$ and $B$.
Notice that right-absorption cannot be expressed as 
\mbox{$\cF(A) = \cF(\Cn(A))$} since $\Cn(A)$ need not be finite.
The best way to study finitary operations is to extend them to infinitary
operations and use the representation theorems we have developed in the
previous section. We shall see that there is a canonical way to extend
finitary operations. But, before we can do that, we must prove one technical
result concerning finitary operations. Its proof is the corresponding part of
the proof of Theorem~\ref{the:repinf}, after replacing
arbitrary sets by finite sets wherever needed.
\begin{lemma}
\label{le:rep}
Let \cF\ be a supraclassical, left-absorbing and deductive finitary operation.
Then, for any finite \mbox{$A \subseteqf \cL$},
\mbox{$\cF(A) = \Cn(A , \bigcap_{B \subseteq A} \cF(B))$}.
\end{lemma}
\QED
In the next section we shall deal with the question of extending
finitary operations.
\section{Co-compactness}
Our notion of compactness is not really satisfying,
since it does not seem to be the {\em right} generalization of the notion
of compactness (for monotonic operations) 
since any finitary monotonic operation
has a unique compact monotonic extension, 
but compact extensions of nonmonotonic operations are not unique.
If we restrict our attention to supraclassical, left-absorbing, deductive
and compact operations the new notion we need seems to be the following.
\begin{definition}
\label{def:co-comp}
An operation \cC\ is said to be {\em strongly co-compact} iff, for any 
\mbox{$X \subseteq \cL$} and for any \mbox{$x \in \cL$},
if \mbox{$x \not \in \cC(X)$}, then
there is a finite \mbox{$A \subseteqf X$} such that
\mbox{$x \not \in \cC(A)$}.
\end{definition}
The following is obvious.
\begin{lemma}
\label{le:co-compact}
If \cS\ is antitonic, it is strongly co-compact iff
\mbox{$\cS(X) = \bigcap_{A \subseteqf X} \cS(A)$}.
\end{lemma}
Not all operations represented by antitonic operators are strongly co-compact,
as may be seen from the following counter-example.
Consider a propositional calculus on an infinite set of propositional
variables, and let \Cn\ be logical consequence.
Let $a$ a proposition that is not a tautology.
Define, for any finite set $B$, 
\mbox{$\cS(B) \eqdef \Cn(\{ \chi_{B} \ra a \})$},
where $\chi_{B}$ is the conjunction of all elements of $B$.
For infinite $X$, \mbox{$\cS(X) \eqdef \bigcap_{B \subseteqf X} \cS(B)$}.
It is easy to see that \mbox{$a \in \cC(B) = \Cn(B , \cS(B))$}, 
for any finite set 
\mbox{$B \subseteqf \cL$}. Let $X$ be any infinite set of propositional
variables that do not appear in $a$. 
One may see that \mbox{$\bigcap_{B \subseteqf X} \cS(B)$}
is equal to $\Cn(\emptyset)$, and therefore \mbox{$\cC(X) = \Cn(X)$}.
We conclude that \mbox{$a \not \in \cC(X)$}, even though
\mbox{$a \in \cC(A)$} for any finite subset $A$ of $X$.
Many operations represented by antitonic operations are strongly co-compact, 
though.
\begin{theorem}
\label{the:finS}
If \mbox{$\cC(X) = \Cn(X , \cS(X))$} for some antitonic and strongly 
co-compact \cS , 
and
if $\cS(\emptyset)$ is finite, then \cC\ is strongly co-compact.
\end{theorem}
\proof
Since \cS\ is antitonic and strongly co-compact, 
we have 
\linebreak[4]
\mbox{$\cC(X) = \Cn(X , \bigcap_{B \subseteqf X} \cS(B))$}.
Since, for any $Y$,
\mbox{$\cS(Y) \subseteq \cS(\emptyset)$} and this last set is finite,
the intersection above is a finite intersection.
By Lemma~\ref{le:IGPL}, part~\ref{adm}, 
\mbox{$\cC(X) = \bigcap_{B \subseteqf X} \Cn(X , \cS(B))$}.
Suppose \mbox{$x \not \in \cC(X)$}. There is a finite set 
\mbox{$B \subseteqf X$}, such that
\mbox{$x \not \in \Cn(X , \cS(B))$}.
Clearly, \mbox{$x \not \in \Cn(B , \cS(B)) = \cC(B)$}.
\QED
Our next result shows that compactness and strong co-compactness
together play the role of a generalization of the notion of compactness.
\begin{theorem}
\label{the:uniqueext}
Let \cF\ be a finitary supraclassical, left-absorbing, deductive operation.
It has a unique supraclassical, left-absorbing, deductive, compact
and strongly co-compact extension.
\end{theorem}
\proof
Suppose \cF\ is a finitary supraclassical, left-absorbing, deductive operation.
We shall prove uniqueness of the extension first.
Suppose that \cC\ is an infinitary supraclassical, left-absorbing, 
deductive, compact and strongly co-compact extension of \cF.
By Theorem~\ref{the:repinf},  
\mbox{$\cC(X) = \Cn(X , \bigcap_{Y \subseteq X} \cC(Y))$}.
But \cC\ is strongly co-compact, and 
\[
\bigcap_{Y \subseteq X} \cC(Y) = \bigcap_{B \subseteqf X} \cC(B).
\]
We conclude that 
\begin{equation}
\label{eq:extdef}
\cC(X) = \Cn(X , \bigcap_{B \subseteqf X} \cF(B)).
\end{equation}
But Equation~(\ref{eq:extdef}) uniquely defines \cC\ in terms of \cF.

To prove existence, we shall show that Equation~(\ref{eq:extdef})
provides an extension of \cF\ with all the required properties.
Lemma~\ref{le:rep} shows that it is indeed an extension of \cF.
We notice that Equation~(\ref{eq:extdef}) provides an antitonic
representation of \cC\ and therefore Theorem~\ref{the:repinf} 
enables us to conclude that \cC\ is 
supraclassical, left-absorbing, deductive and compact.
It is left to us to show that \cC\ is strongly co-compact.
But this follows clearly from Equation~(\ref{eq:extdef}) since
\mbox{$\bigcap_{B \subseteqf X} \cF(B) \subseteq \cC(X)$}.
\QED
\begin{corollary}
\label{co:AB}
If \cF\ is supraclassical, left-absorbing and deductive, its unique
supraclassical, left-absorbing, deductive, compact and strongly co-compact
extension is given by
\mbox{$x \in \cC(X)$} iff there is a finite \mbox{$A_{x} \subseteqf X$}
such that \mbox{$x \in \cF(A_{x} , B)$} for any finite \mbox{$B \subseteqf X$}.
\end{corollary}
\proof
Let $\cC'$ be the unique extension of \cF\ 
provided by Equation~(\ref{eq:extdef}).
One easily checks that \cC\ is also an extension of \cF, and, since $\cC'$ is
compact, we have \mbox{$\cC'(X) \subseteq \cC(X)$}.
Suppose now that \mbox{$x \in \cC(X)$}.
We have \mbox{$x \in \bigcap_{B \subseteqf X} \cF(A_{x} , B)$}.
Since \cF\ is deductive, by Lemma~\ref{le:IGPL} and 
Equation~(\ref{eq:extdef}),
\[
x \in \bigcap_{B \subseteqf X} \Cn(A_{x} , B , \cF(B))
= \bigcap_{B \subseteqf X} \Cn(A_{x} , \cF(B)) =
\Cn(A_{x} , \bigcap_{B \subseteqf X} \cF(B)) 
\]
\[
\subseteq \Cn(X , \bigcap_{B \subseteqf X} \cF(B)) =
\cC'(X).
\]
\QED
\section{Right-absorbing operations}
We shall now deal with extending right-absorbing finitary 
operations.
It turns out that we cannot prove that any supraclassical, left-absorbing, 
right-absorbing and deductive finitary operation has a supraclassical, 
left-absorbing, right-absorbing, deductive, compact and
strongly co-compact extension. 
We, therefore, need a weaker notion of co-compactness.
\begin{definition}
\label{def:wcocomp}
An operation \cC\ is said to be {\em co-compact} iff, for any 
\mbox{$X \subseteq \cL$} and for any \mbox{$x \in \cL$},
if \mbox{$x \not \in \cC(X)$}, then
there is a finite set $A$, such that \mbox{$X \models A$} and
\mbox{$x \not \in \cC(A)$}.
\end{definition}
Clearly, any strongly co-compact operation is co-compact.
We also have the following, the proof of which is obvious.
\begin{lemma}
\label{le:CCn}
If \cC\ is strongly co-compact, then the operation $\cC'$ defined by
\linebreak[4]
\mbox{$\cC'(X) = \cC(\Cn(X))$} is co-compact.
\end{lemma}
We may now prove our result concerning right-absorbing operations.
\begin{theorem}
\label{the:uniqwco}
Any supraclassical, left-absorbing, right-absorbing and deductive
finitary operation \cF\ has a unique extension that is supraclassical, 
left-absorbing, right-absorbing, deductive, compact and co-compact.
\end{theorem}
\proof
Let \cC\ be the unique supraclassical, left-absorbing, deductive,
compact and strongly co-compact extension of \cF\ 
defined by Equation~(\ref{eq:extdef}).
Consider $\cC'$ defined by \mbox{$\cC'(X) = \cC(\Cn(X))$}.
The operation $\cC'$ is easily seen to be  supraclassical, left-absorbing, 
right-absorbing, deductive and compact. 
By Lemma~\ref{le:CCn}, it is also co-compact.
By Corollary~\ref{co:AB},
\mbox{$x \in \cC'(X)$} iff there is a finite \mbox{$A_{x} \subseteqf \Cn(X)$}
such that \mbox{$x \in \cF(A_{x} , B)$} for any finite 
\mbox{$B \subseteqf \Cn(X)$}.
This shows immediately that $\cC'$ is an extension of \cF.
We have proved existence.

For uniqueness notice that, if \cC\ has the properties required,
we have 
\linebreak[4]
\mbox{$\cC(X) = \Cn(X , \bigcap_{X \models Y} \cC(Y))$},
by Theorem~\ref{the:repinfra}.
But, if \cC\ is a co-compact extension of \cF,
\linebreak[3]
\mbox{$\bigcap_{X \models Y} \cC(Y) = \bigcap_{B \subseteqf \Cn(X)} \cC(B)$}.
\QED
The extension of Theorem~\ref{the:uniqwco} is characterized in the
following way.
\begin{corollary}
\label{co:AB2}
If \cF\ is supraclassical, left-absorbing, right-absorbing and deductive, 
its unique
supraclassical, left-absorbing, right-absorbing, deductive, 
compact and co-compact
extension is given by
\mbox{$x \in \cC(X)$} iff there is a finite \mbox{$A_{x} \subseteqf X$}
such that \mbox{$x \in \cF(A_{x} , B)$} for any finite 
\mbox{$B \subseteqf \Cn(X)$}.
\end{corollary}
\proof
In the proof of Theorem~\ref{the:uniqwco} we have seen this extension
is $\cC(\Cn(X))$ for the extension \cC\ that is described in 
Corollary~\ref{co:AB}.
Therefore \mbox{$x \in \cC(X)$} iff there is a finite 
\mbox{$A_{x} \subseteqf \Cn(X)$}
such that \mbox{$x \in \cF(A_{x} , B)$} for any finite 
\mbox{$B \subseteqf \Cn(X)$}.
But if \mbox{$A_{x} \subseteqf \Cn(X)$}, there is a finite 
\mbox{$A'_{x} \subseteqf X$} such that \mbox{$A'_{x} \models A_{x}$}, 
and therefore
\linebreak[3]
\mbox{$\cF(A'_{x} , B) = \cF(A'_{x} , A_{x} , B)$} by right-absorption.
\QED
Corollary~\ref{co:AB2} shows that the unique extension of \cF\ is the 
operation denoted $\cC_{\cF}$ in~\cite{FL:IGPL} (the only case considered
there was the case of a cumulative \cF).
Our next result deals with cumulative operations.
\begin{theorem}
\label{the:cumuni}
The unique supraclassical, left-absorbing, right-absorbing, deductive,
compact and co-compact extension \cC\ of a supraclassical,
left-absorbing, right-absorbing, deductive and cumulative finitary
operation \cF\ is cumulative. 
\end{theorem}
\proof
We have seen in the proof of Theorem~\ref{the:uniqwco} that this 
unique extension is defined by
\mbox{$\cC(X) = \Cn(X , \bigcap_{X \models B} \cF(B))$}.
By the first leg of Theorem~\ref{the:newrep}, all we have to show is that
\mbox{$\cS(X) \eqdef \bigcap_{X \models B} \cF(B)$} is cumulative.
We immediately see that \cS\ is antitonic.
Since \cF\ is left-absorbing, so is \cS.
Suppose, then, that \mbox{$Y \subseteq \cS(X)$}.
We must show that \mbox{$\cS(X) = \cS(X , Y)$}.
By antitonicity of $\cS$, we immediately see that 
\mbox{$\cS(X , Y) \subseteq \cS(X)$}.
We shall show that \mbox{$\cS(X) \subseteq \cS(X , Y)$}.
Let $A$ be an arbitrary finite set such that \mbox{$X , Y \models A$},
we shall show that \mbox{$\cS(X) \subseteq \cF(A)$}.
Since \mbox{$X , Y \models A$}, for any \mbox{$a \in A$},
there is a finite \mbox{$Y_{a} \subseteqf Y$}, such that 
\mbox{$X \models Y_{a} \ra a$}. In fact, \mbox{$Y_{a} \ra a$} is a
singleton.
Let \mbox{$B \eqdef \bigcup_{a \in A } Y_{a} \ra a$}.
The set $B$ is finite.
Notice that \mbox{$A \models B$} and \mbox{$B , Y \models A$}.
There is therefore a finite subset $C$ of $Y$ such that 
\mbox{$B , C \models A$}.
We see that \mbox{$\Cn(B , C) = \Cn(A , C)$}.
Since \mbox{$X \models B$}, we have, by the definition of \cS, 
\mbox{$\cS(X) \subseteq \cF(B)$}.
It will be enough to prove that \mbox{$\cF(B) = \cF(A)$}.
But, indeed, \mbox{$C \subseteqf Y \subseteq \cS(X) \subseteq \cF(B)$} 
and by the cumulativity of \cF,
\mbox{$\cF(B) = \cF(B , C)$}.
By right-absorption \mbox{$\cF(B , C) = \cF(A , C)$}.
We conclude that \mbox{$A \subseteqf \cF(B)$}, and, by cumulativity of \cF,
\mbox{$\cF(B) = \cF(B , A)$}, but \mbox{$\Cn(B , A) = \Cn(A)$},
and we conclude that \mbox{$\cF(B) = \cF(A)$}.
\QED

Theorem~\ref{the:cumuni} represents an improvement on Theorem 7.9 
of~\cite{FL:IGPL}, in which the language \cL\ was assumed to have a
disjunction.
There, it was shown that, under the additional assumption that the language
\cL\ has a disjunction, the ``canonical'' extension of a supraclassical,
left-absorbing, right-absorbing, deductive and cumulative finitary operation
\cF\ is the $\cC_{\cF}$ described in Corollary~\ref{co:AB2}, and is
supraclassical, left-absorbing, right-absorbing, deductive and cumulative.
This operation $\cC_{\cF}$ is, trivially, compact and co-compact.
\section{Conclusion, open problems and acknowledgments}
In our introduction we mentioned a number of nonmonotonic systems.
We may now try to tell which ones have antitonic representations.
As mentioned in the introduction, the Closed World Assumption 
of~\cite{Rei:CWA} is explicitly presented by an antitonic representation.
Minker's~\cite{Mink:GCWA} Generalized Closed world Assumption, 
on the contrary, does not have
any such representation since the inference operation it defines does not 
satisfy Deductivity.
Indeed from the assumptions \mbox{$\{ p , p \vee q \}$}, 
GCWA will conclude $\neg q$, but
from $p \vee q$ it will not conclude \mbox{$p \ra \neg q$}.
Notice that, equivalently, GCWA does not satisfy Or since it will
conclude $\neg q$ from $p$ and conclude $\neg p$ from $q$, but will not
conclude \mbox{$\neg p \vee \neg q$} from \mbox{$p \vee q$}.

Default Logic~\cite{Reiter:80} is a bit more problematic to study since it
does not explicitly defines an inference operation. If we take the reasonable,
skeptical approach to this definition, we see that, even when only normal
defaults are considered, it lacks an antitonic representation, since its 
inference operation is not deductive (or does not satisfy Or).
The example given above for GCWA translates immediately in normal default
logic.
It is all the more remarkable that, if we restrict ourselves to finite sets
of normal
defaults without prerequisites, the skeptical inference operation defined
is deductive and admits an antitonic representation.
Indeed, normal defaults without prerequisites are equivalent to Poole
systems without constraints.
In~\cite{FL:IGPL}, Theorems 7.17 and previous theorems, it was shown that, 
if the language \cL\
has a contradiction (i.e. if any inconsistent set has a finite subset that
is inconsistent), the inference operation defined by any finite Poole system
without constraints satisfies the conditions of Theorem~\ref{the:newrep},
part~\ref{supracu}.
For Circumscription, since the inference operation defined by 
Circumscription may be defined by some preferential model, 
it is clearly supraclassical, right-absorbing, left-absorbing and deductive. 
When the so-called
{\em well-foundedness} assumption is satisfied, it is also cumulative.
If the language is logically finite, it is also, obviously, compact and 
therefore has an antitonic representation by Theorem~\ref{the:newrep}.
We do not know yet whether it is compact even when the language is infinite.
It follows from~\cite{LMAI:92} Section 5.8 that,
when the knowledge base is admissible,
rational closure has an antitonic representation.

We want to thank Michael Freund for helping us to show that cumulative
compact operations are supracompact and David Makinson, Michael Freund
and three anonymous referees for their remarks on a draft of 
this paper.


\begin{thebibliography}{10}

\bibitem{Brass:93}
Stefan Brass.
\newblock On the semantics of supernormal defaults.
\newblock In Ruzena Bajcsy, editor, {\em Proceedings of the 13th I.J.C.A.I.},
  pages 578--583. Morgan Kaufmann, Chamb\'{e}ry, Savoie, France, August 1993.

\bibitem{Freund:supra93}
Michael Freund.
\newblock Supracompact inference operations.
\newblock {\em Studia Logica}, 52:457--481, 1993.

\bibitem{FL:IGPL}
Michael Freund and Daniel Lehmann.
\newblock Nonmonotonic inference operations.
\newblock {\em Bulletin of the IGPL}, 1(1):23--68, July 1993.
\newblock Produced by the Max-Planck-Institut f\"{u}r Informatik, Im Stadtwald,
  D-66123 Saarbr\"{u}cken, Germany.

\bibitem{GarMak:92}
Peter G\"{a}rdenfors and David Makinson.
\newblock Nonmonotonic inference based on expectations.
\newblock {\em Artificial Intelligence}, 65(1), January 1994.

\bibitem{KLMAI:89}
Sarit Kraus, Daniel Lehmann, and Menachem Magidor.
\newblock Nonmonotonic reasoning, preferential models and cumulative logics.
\newblock {\em Artificial Intelligence}, 44(1--2):167--207, July 1990.

\bibitem{LMAI:92}
Daniel Lehmann and Menachem Magidor.
\newblock What does a conditional knowledge base entail?
\newblock {\em Artificial Intelligence}, 55(1):1--60, May 1992.

\bibitem{McCarthy:80}
John McCarthy.
\newblock Circumscription, a form of non monotonic reasoning.
\newblock {\em Artificial Intelligence}, 13:27--39, 1980.

\bibitem{Mink:GCWA}
Jack Minker.
\newblock On indefinite databases and the closed world assumption.
\newblock In D.W. Loveland, editor, {\em 6th Conference on Automated
  Deduction}, volume 138 of {\em Lecture Notes in Computer Science}, pages
  292--308. Springer-Verlag, New York, June 1982.

\bibitem{Poole:88}
David Poole.
\newblock A logical framework for default reasoning.
\newblock {\em Artificial Intelligence}, 36:27--47, 1988.

\bibitem{Rei:CWA}
Raymond Reiter.
\newblock On closed world data bases.
\newblock In H.~Gallaire and J.~Minker, editors, {\em Logic and Data Bases},
  pages 55--76. Plenum, New York / London, 1978.

\bibitem{Reiter:80}
Raymond Reiter.
\newblock A logic for default reasoning.
\newblock {\em Artificial Intelligence}, 13:81--132, 1980.

\end{thebibliography}
\end{document}